\documentclass[letterpaper, 10 pt, conference]{ieeeconf}  

\IEEEoverridecommandlockouts                              

\usepackage{balance}
\usepackage{graphicx}
\usepackage{amsmath}
\usepackage{amsfonts}
\usepackage{comment}
\usepackage{booktabs}
\usepackage{xcolor}

\title{\LARGE \bf
Evaluating Human and LLM‑Generated Thematic Analysis in HRI for Vulnerable Populations: A Comparative and Ethical Analysis \vspace{-0.5em}}

\author{Alva Markelius$^{1, *}$, Fethiye Irmak Dogan$^{1, *}$, Julie Bailey$^{2}$ and Hatice Gunes$^{1}$\vspace{-0.5em}
\thanks{$^{1}$Affective Intelligence and Robotics (AFAR) Lab, Department of Computer Science and Technology, University of Cambridge, UK.
$^{2}$Faculty of Education, University of Cambridge, UK.
        {\tt\small \{ajkm4, fid21, jb658, hg410\}@cam.ac.uk}}%
\thanks{* Both authors contributed equally to this research.}
\thanks{AM is supported by Cambridge Trust. FID and HG are supported in part by CHANSE and NORFACE through the MICRO project, funded by UKRI/ESRC (grant ref. UKRI572).}
}

\begin{document}

\maketitle
\thispagestyle{empty}
\pagestyle{empty}

\begin{abstract}
Thematic analysis (TA) has long been regarded as an inherently human, reflexive, and interpretive process. However, the extent to which LLM-generated TA is appropriate for Human-Robot Interaction (HRI) research involving vulnerable populations remains largely unexamined and raises critical questions about validity and ethics, particularly in sensitive research contexts. This paper presents a comparative study of human- and LLM-generated TA in an HRI context with a focus on vulnerable populations. We evaluate both objective and semantic agreement between human- and LLM-generated themes, and examine whether observed divergences reflect systematic interpretive patterns with ethical significance. Our analysis investigates whether LLM-generated TA risks marginalising or misrepresenting the experiences of vulnerable participants, with implications for researchers employing LLM-assisted TA in HRI.\end{abstract}

\section{Introduction}
Thematic analysis (TA) has long been understood as an inherently human interpretive process that demands contextual judgment, reflexive engagement with data, and is shaped by the researcher's subjectivity \cite{terry2017thematic}, positionality \cite{Pascoe_2022}, domain knowledge, cultural awareness, and ethical sensibility \cite{Braun_Clarke_2022}. Indeed, coding qualitative data is itself a complex craft of constructing meaning  \cite{williams2019art}. 

Recently, researchers have begun to explore Large Language Models (LLMs) as tools for coding and theme generation \cite{Ornelas_Arajo_Ara_Trinkenreich_Kalinowski_2025, Montes_Feldt_Martos_Ouhbi_Premanandan_Graziotin_2025}, reporting promising levels of agreement with human analysts \cite{Castellanos_Jiang_Gomes_VanderMeer_Castillo_2025}. However, the extent to which these findings generalise to sensitive domains, particularly Human-Robot Interaction (HRI) research involving vulnerable populations such as people with disabilities or children, remains largely unexamined. This gap is consequential since vulnerable populations introduce unique interpretive demands: lived experience, marginalised perspectives, and domain-specific contextual knowledge that may not be adequately captured by general-purpose LLMs. Furthermore, existing comparative studies have primarily evaluated LLM-generated TA through inter-rater agreement metrics alone \cite{Mathis_Zhao_Pratt_Weleff_DePaoli_2024, Parkington_Teferra_RouleauTang_Perivolaris_Rueda_Dubrowski_Kapralos_Samavi_Greenshaw_Zhang_etal._2025}, without systematically examining the ethical nature or directionality of disagreements. Indeed, when discrepancies arise, it is unknown whether they are random or systematic and whether systematic disagreements carry ethical implications for how findings are constructed and whose experiences are centred.

To address these limitations, we present a comparative study of human- and LLM-generated TA applied to interview data from a user study involving 31 university students with disabilities, conducted in the context of an HRI application \cite{10.1145/3757279.3785593}. To the authors' knowledge, this is \textit{the first comparative analysis investigating the feasibility and appropriateness of LLM-based TA in an HRI context, and in particular with a focus on vulnerable populations}. While the empirical data are drawn exclusively from participants with disabilities, this population serves as an illustrative case and the analytical findings and ethical implications are developed from the broader perspective of vulnerable populations in HRI, and are intended to generalise across groups whose characteristics may introduce heightened methodological and ethical risks.


Beyond measuring agreement, we analyse the character of disagreements and examine whether observed divergences reflect systematic patterns with ethical significance for HRI research with vulnerable groups. Specifically, we ask: (RQ1) to what extent do human and LLM analyses assign utterances to the same thematic groupings, evaluated at the theme and subtheme levels; (RQ2) how semantically similar are the human- and LLM-generated labels assigned to the same utterance, evaluated at the code, subtheme, and theme levels; and (RQ3) to what extent do disagreements reflect systematic interpretive differences rather than random variation, examined through an ethical lens with particular attention to whether divergences risk marginalising or misrepresenting the experiences of vulnerable or marginalised participants.

\section{Related Work}
\subsection{LLMs for Thematic Analysis}
The use of LLMs to automate or augment TA raises methodological, epistemological, and ethical questions of interpretive authority, bias, and transparency \cite{Schroeder_AubinRandazzo_Mimno_Schoenebeck_2025}. Embedding LLMs within established frameworks such as Braun and Clarke's six-phase method \cite{braun2021thematic} reports efficiency gains in open coding and theme generation, but also risks of contextual oversimplification, bias amplification, and premature interpretive closure \cite{Ornelas_Arajo_Ara_Trinkenreich_Kalinowski_2025}, indicating that human validation and reflexive oversight remain indispensable. Although empirical comparisons report substantial agreement (approximately 80\%) between LLM- and human-generated themes \cite{Castellanos_Jiang_Gomes_VanderMeer_Castillo_2025}, discrepancies emerge where subtle contextual or domain-specific interpretation is required, with LLMs performing better on descriptive, surface-level themes than on deeply interpretive or culturally embedded ones \cite{Castellanos_Jiang_Gomes_VanderMeer_Castillo_2025}. Comparing coding tasks, \cite{Wen_Clough_Paton_Middleton_2026} finds substantial agreement for deductive themes, but for inductive themes a tendency toward overgeneralisation despite strong semantic similarity. Practical challenges also remain, including hallucination, variability across model versions, instability in excerpt extraction, and the interpretive labour required to supervise, refine, and ethically govern LLM-assisted TA \cite{Wen_Clough_Paton_Middleton_2026}.

Existing work has focused largely on general qualitative datasets, with less attention to the ethical implications of LLM-assisted TA in research involving vulnerable populations, where misinterpretation, flattening of lived experience, or amplification of normative biases carry heightened ethical stakes. Evaluations have also predominantly assessed aggregate agreement, rather than whether systematic human-LLM differences reflect deeper epistemic or normative divergences. We move beyond performance evaluation alone to investigate (i) the overall consistency between human- and LLM-generated themes across an HRI study, and (ii) whether observed divergences are systematic in ways that raise ethical concerns when analysing data from vulnerable groups.

\subsection{Thematic Analysis in HRI for Vulnerable Populations}
Qualitative methods (without LLMs), including e.g. interviews, focus groups, and textual analysis, are widely employed in HRI research \cite{Veling_McGinn_2021}. TA is considered particularly critical when working with vulnerable groups, as they can empower participants "to engage with the technology from a position of authority, and to be in a position to challenge the norms, perceptions and biases" inherent to the field \cite{Veling_McGinn_2021}. TA, as a structured qualitative method, has consequently become a common analytical tool across HRI studies involving vulnerable and often marginalised communities.

In paediatric contexts, TA has evaluated robot-assisted speech therapy for children with language disorders \cite{Estvez_Terano_2021} and telepresence robots for children with chronic illness \cite{Page_Charteris_Berman_2021}. In care settings, it has surfaced perspectives of staff in long-term dementia care \cite{Moyle_Bramble_Jones_Murfield_2018} and what older adults seek from social robot companions \cite{Sraa_Tndel_Kharas_Serrano_2023}. Beyond age-related vulnerability, TA has informed the deployment of humanoid robots for traditional language learning in indigenous classrooms \cite{Keane_andWilliams_2019}, examined teacher perceptions of robots for children with migration backgrounds \cite{Tozadore_Guneysui_2023}, and a platform supporting refugee integration \cite{Simo_Avelino_Duarte_Figueiredo_2018}. It has also assessed socially assistive robots among LGBT+ and ethnic minority youth at risk of self-harm \cite{Williams_Townsend_Naeche_Chapman-Nisar_Hollis_Slovak_Minds_2023}, and the values of LGBT+ older adults regarding robots designed to alleviate loneliness \cite{Poulsen_Burmeister_Greig_Ulhaq_Tien_2025}.

These studies establish TA as an ethically motivated methodology for capturing lived experiences of vulnerable populations in HRI. As this work begins incorporating LLMs, scrutinising the analytical process itself becomes equally important. We address this by evaluating not only the consistency of LLM-generated TA against human analysis in HRI, but also whether any divergences carry ethical weight when the data originates from vulnerable populations.

\section{Methodology}
\vspace{-0.3em}
This study uses qualitative data collected in a prior within-subjects HRI user study examining the use of social robots and voice agents as mediation support tools for disabled university students~\cite{10.1145/3757279.3785593}. The original study employed a 2$\times$2 within-subjects design ($N = 31$; $M_{age} = 25.2$, $SD = 6.7$) crossing two support roles---\textit{Signposting} (informational guidance to university resources) and \textit{Sounding Board} (facilitated self-reflection and problem-solving)---with two agent embodiments: a Pepper robot using a hybrid GPT-4/scripted dialogue system, and an Amazon Echo Dot running identical software. Participants represented (self-reported) diverse disability profiles, including Autism ($n = 5$), Specific Learning Differences ($n = 7$), and Mental Health conditions ($n = 7$). Each participant completed a semi-structured interview about their perception of robot/agent based disability support, which was audio-recorded and subsequently transcribed. Details of the experimental design, participant recruitment, robotic platforms, and dialogue architecture are reported in~\cite{10.1145/3757279.3785593}. 

\subsection{Human Thematic Analysis}
\vspace{-0.3em}
A human researcher, with expertise in disability and neurodivergence in educational settings, conducted reflexive thematic analysis of the interaction transcripts following Braun and Clarke's six-phase framework~\cite{braun2021thematic}. Familiarisation with the data was achieved through active engagement with the corpus during transcription editing and video review. Following familiarisation, the researcher identified salient segments across the transcripts and extracted these into a dedicated analysis document; coding was conducted on these extracted segments rather than on the raw transcripts, in order to preserve transcript integrity and improve tractability. Initial codes were generated inductively, then progressively collated and reviewed across the full dataset to construct candidate themes. Themes were subsequently reviewed, refined, defined, and named in accordance with the framework, with the final thematic structure reflecting participants' perspectives on their interactions across the four conditions.

\subsection{LLM Thematic Analysis}
\vspace{-0.3em}

For the LLM thematic analysis, we synthesised and adapted prompting strategies from prior work~\cite{Naeem_Smith_Thomas_2025, Ornelas_Arajo_Ara_Trinkenreich_Kalinowski_2025}, following Braun and Clarke's six-phase framework~\cite{braun2021thematic} to ensure methodological consistency across both analyses. A system prompt assigned the model (Claude Sonnet 4.6~\cite{anthropic2025system} with extended thinking) the role of thematic analyst and presented each phase as a separate, sequential task. Beyond the transcripts, the model was provided with domain-specific context: the study's purpose and $2\times2$ design, a description of the participant population and interaction conditions, definitions of each phase of the reflexive method, and the analytic focus on participants' perspectives across the robot and agent conditions. Because the corpus comprised 31 participants, Phases 1 and 2 (familiarisation and initial code generation) were run in four batches to mitigate degraded output quality under high token load. Codes from all four batches were then pooled into a single codebook, and Phases 3 to 6 (generating, reviewing, defining, and naming themes) were conducted over this consolidated set across the full corpus rather than per batch. This pooled reanalysis served as a cross-batch normalisation step, collapsing redundant codes generated for similar phenomena across batches into shared themes. All prompts are provided in our open repository\footnote{https://github.com/AlvaMarkelius/LLM-TA-Paper-Artifacts}.

\subsection{Clustering Analysis}
\vspace{-0.3em}


To evaluate whether human and LLM annotations grouped utterances in a similar manner (RQ1), we conducted a clustering consistency analysis. Importantly, this analysis does not consider the semantic content of labels, but instead evaluates whether the same utterances are assigned to the same groups (theme or subtheme levels) across annotators. Each utterance $i$ is associated with two categorical assignments: a human label $u_i$ and an LLM label $v_i$, coming either from theme or subtheme level labelings. These assignments define human or LLM clusterings over the same set of utterances. To quantify the similarity between these clusterings, we use \textit{Adjusted Mutual Information} (AMI)~\cite{JMLR:v11:vinh10a}, which measures the agreement between clusterings while correcting for chance. Formally, given two clusterings $U$ (human) and $V$ (LLM), the AMI is defined as:
\vspace{-0.2em}
\begin{equation}
\text{AMI}(U, V) = \frac{\text{MI}(U, V) - \mathbb{E}[\text{MI}(U, V)]}{\max\{H(U), H(V)\} - \mathbb{E}[\text{MI}(U, V)]},
\end{equation}%
\noindent where $\text{MI}(U, V)$ is the mutual information between the two clusterings, $H(U)$ and $H(V)$ are their entropies, and $\mathbb{E}[\text{MI}(U, V)]$ is the expected mutual information under random assignment. An AMI value of $1$ indicates perfect agreement, $0$ indicates agreement at the chance level, and negative values indicate less agreement than expected by chance. In our implementation, we used the \texttt{sklearn} library to compute AMI, and we calculated it separately for theme-level and subtheme-level, yielding an observed AMI score in each level.

\subsection{Semantic Analysis}
\vspace{-0.3em}
To evaluate the semantic alignment between human and LLM annotations (RQ2), we computed the semantic similarity between corresponding labels (codes, subthemes, and themes) for each utterance. Each label (e.g., a human code and its corresponding LLM code) was embedded into a continuous vector space using a pretrained sentence embedding model\footnote{https://huggingface.co/sentence-transformers/all-mpnet-base-v2} (\texttt{all-mpnet-base-v2}). Let $\mathbf{h}_i$ and $\mathbf{l}_i$ denote the embedding vectors for the human and LLM labels of utterance $i$, respectively. Semantic similarity was then computed using cosine similarity:
\vspace{-0.2em}
\begin{equation}
\text{cos}(\mathbf{h}_i, \mathbf{l}_i) = \frac{\mathbf{h}_i \cdot \mathbf{l}_i}{\|\mathbf{h}_i\| \|\mathbf{l}_i\|}.
\end{equation}

Since embeddings were normalized, this reduces to the dot product between vectors. This computation was performed separately for codes, subthemes, and themes. For each level, we reported the mean cosine similarity and standard deviation across all utterances per human-provided theme:
\vspace{-0.2em}
\begin{equation}
\begin{aligned}
\mu = \frac{1}{N} \sum_{i=1}^{N} \text{cos}(\mathbf{h}_i, \mathbf{l}_i),\\
\sigma = \sqrt{\frac{1}{N} \sum_{i=1}^{N} (\text{cos}(\mathbf{h}_i, \mathbf{l}_i) - \mu)^2},
\end{aligned}
\end{equation}
\noindent where $N$ shows the total number of samples in each human theme. This provides a measure of both central tendency and variability in semantic agreement.


\subsection{Qualitative Analysis}
\vspace{-0.3em}
To answer RQ3, we selected 3 utterances with the highest and lowest cosine similarity for codes and conducted a qualitative comparison of human and LLM-generated codes. This sampling was chosen because low-similarity captures the most pronounced interpretive divergences, while high-similarity provides a baseline against which to assess whether surface-level agreement masks subtler representational differences. We decided to focus on codes, as they are considered one of the most complex and fundamental aspects of TA~\cite{williams2019art} and are more specific to each utterance than themes and subthemes. To evaluate the comparisons, we drew on literature aimed at identifying tensions, limitations, and ethical implications of LLM-based TA. 
More specifically, we examined the tensions surrounding the use of LLMs in qualitative analysis~\cite{Schroeder_AubinRandazzo_Mimno_Schoenebeck_2025} and considered the limitations noted in~\cite{Ornelas_Arajo_Ara_Trinkenreich_Kalinowski_2025} to formulate a set of qualitative evaluation criteria. From these sources, we derived three qualitative evaluation criteria applied to each sampled case. \textbf{Interpretive Grounding} assesses whether the LLM code constitutes a genuine interpretive label for the utterance or merely a paraphrase of its surface content: a \textit{High} rating indicates close engagement with the data and tight grounding in its meaning, while \textit{Low} indicates extractive or pattern-generalised coding that does not sit with the utterance in depth. \textbf{Representational Fidelity} assesses the degree to which the LLM code accurately reflects the meaning as expressed by the participant: a \textit{High} rating indicates faithful capture of participant voice, while \textit{Low} indicates the code reframes or displaces the participant's expressed meaning. \textbf{Bias and Marginalisation Risk} assesses model bias and the ethical consequence for vulnerable participants: a \textit{High} rating indicates the code is plausibly shaped by biased model properties and risks silencing or misrepresenting a participant's experience, while \textit{Low} indicates plausibly no bias and minimal risk of harm to participant representation. Each criterion was rated as \textit{Low}, \textit{Medium}, or \textit{High} by one researcher (first author, female, 27 y/o) with higher university training and research experience in qualitative methods and AI ethics, based on close reading of each utterance alongside its human and LLM-generated codes. Ratings were assigned through interpretive judgment grounded in the criteria definitions, with written rationales recorded and provided for each case to ensure transparency and replicability.

\section{Findings}
\vspace{-0.3em}

\subsection{Human and LLM themes}
\vspace{-0.3em}
This section briefly explains the final themes derived by the human/LLM to enable better understanding and context of the comparison. The final five human themes were: \textit{Fluency of interaction} meaning is the participant being heard and understood, \textit{Effort requirements} relating to how much social effort does the interaction take, \textit{Sensitivity to disability needs} relating to if the participant's disability needs are understood, \textit{Power dynamics} relating to being in control and empowerment and finally \textit{Social attributes} interrogating if it is a satisfying interaction. The final LLM themes were \textit{Comparing Interaction Modes}, \textit{Experiencing Psychological Safety and Reduced Social Burden}, \textit{Navigating the Robot's Embodied Presence}, \textit{Negotiating Risks, Limitations, and Social Concerns}, and \textit{Recognising Practical Value for Disability Navigation and Advocacy}.

\subsection{RQ1: Clustering Consistency}
\vspace{-0.3em}
\begin{figure}[t!]
    \centering
    \includegraphics[width=1\linewidth]{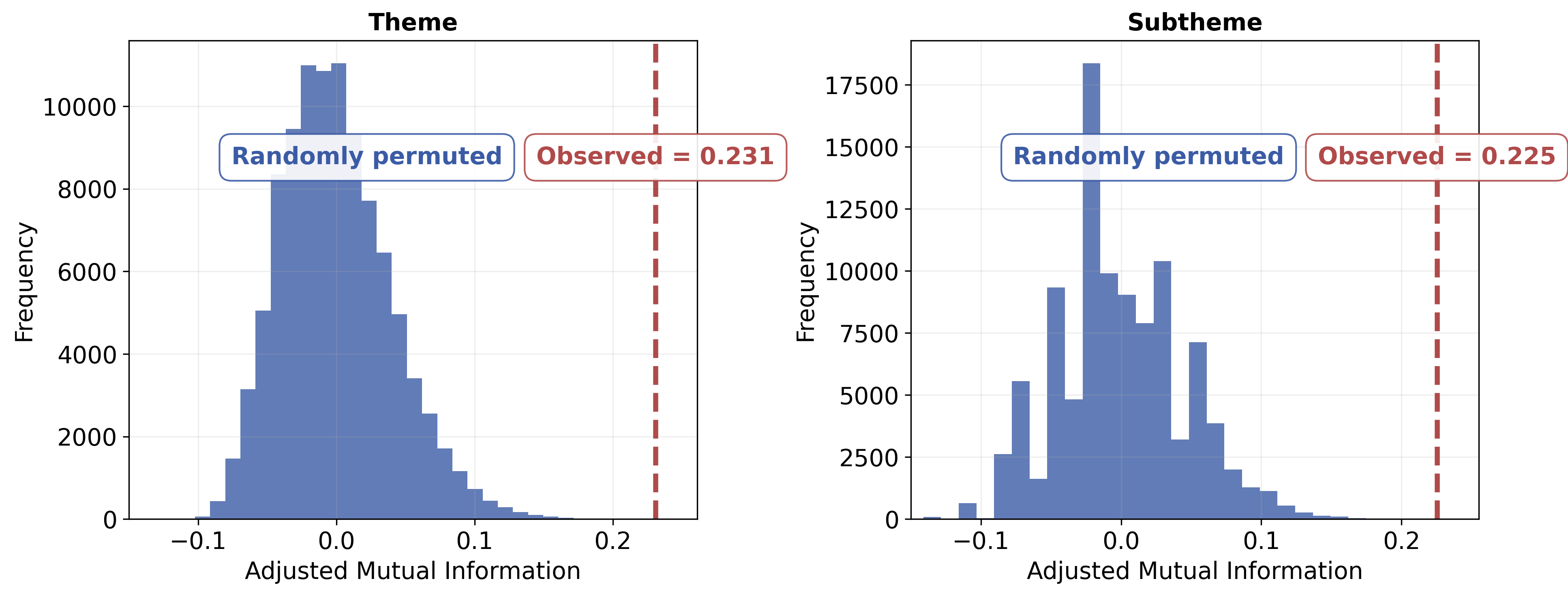}
    \vspace{-2em}
    \caption{Clustering consistency of human and LLM thematic analysis.}
    \label{fig:RQ1}
    \vspace{-0.5em}
\end{figure}

\begin{figure}[t!]
    \centering
    \includegraphics[width=0.8\linewidth]{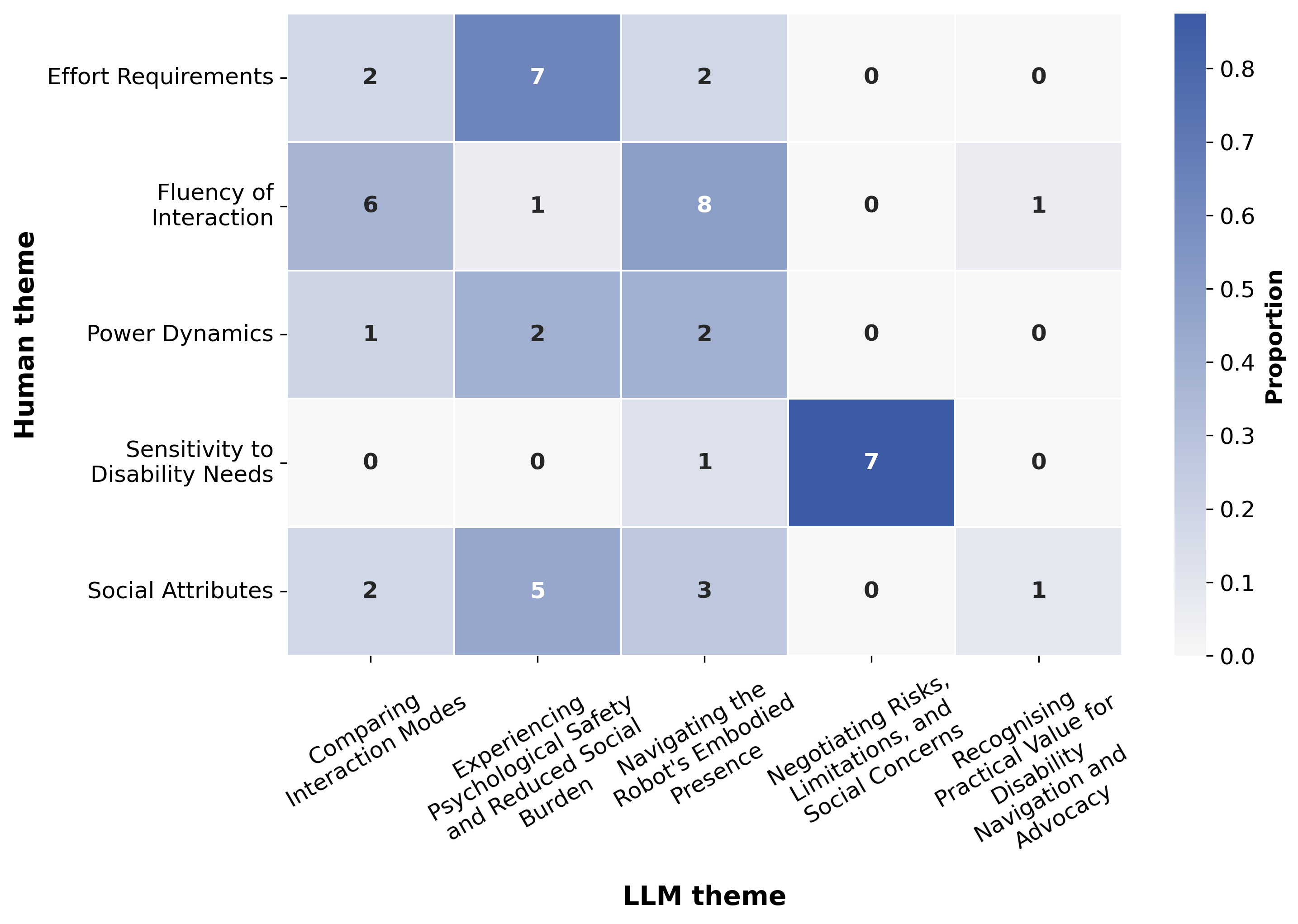}
    \vspace{-1em}
    \caption{Confusion matrix between human and LLM themes.}
    \label{fig:RQ1_1}
    \vspace{-2em}
\end{figure}


To assess the consistency of human and LLM clusterings, we first generated a null distribution by randomly permuting the LLM labels 100{,}000 times and recomputing AMI for each permutation. This produces an empirical distribution of AMI values expected under random assignment as shown in Figure~\ref{fig:RQ1}. As expected, the null distribution is centered around zero, with some negative values. Negative AMI values arise due to the adjustment for chance, indicating that some random permutations produce clusterings that are less aligned than expected by chance alone. On the other hand, the observed AMI values were overall higher than this null distribution, as shown in Figure~\ref{fig:RQ1}. At the theme level, the observed AMI was $0.231$, and at the subtheme level, it was $0.225$. Both values are well above chance, indicating that the LLM produced clusters that align with human clusters beyond chance level. 
To better understand how themes align, Figure~\ref{fig:RQ1_1} presents a confusion matrix between human and LLM theme assignments. The strongest correspondence is observed between \textit{``sensitivity to disability needs''} and \textit{``negotiating risks, limitations and social concerns''}. This is followed by a strong alignment between \textit{``effort requirements''} and \textit{``experiencing psychological safety and reduced social burden''}. These patterns indicate that, despite differences in the wording of themes, the LLM can capture structurally similar groupings of utterances across multiple themes. In contrast, the human theme \textit{``power dynamics''} is distributed relatively evenly across three LLM themes, indicating weaker alignment. This suggests that certain themes are more ambiguous or conceptually diffuse, making them harder for the LLM to consistently map to a single coherent category.

\subsection{RQ2: Semantic Consistency}

\begin{figure}[t!]
    \centering
    \includegraphics[width=0.8\linewidth]{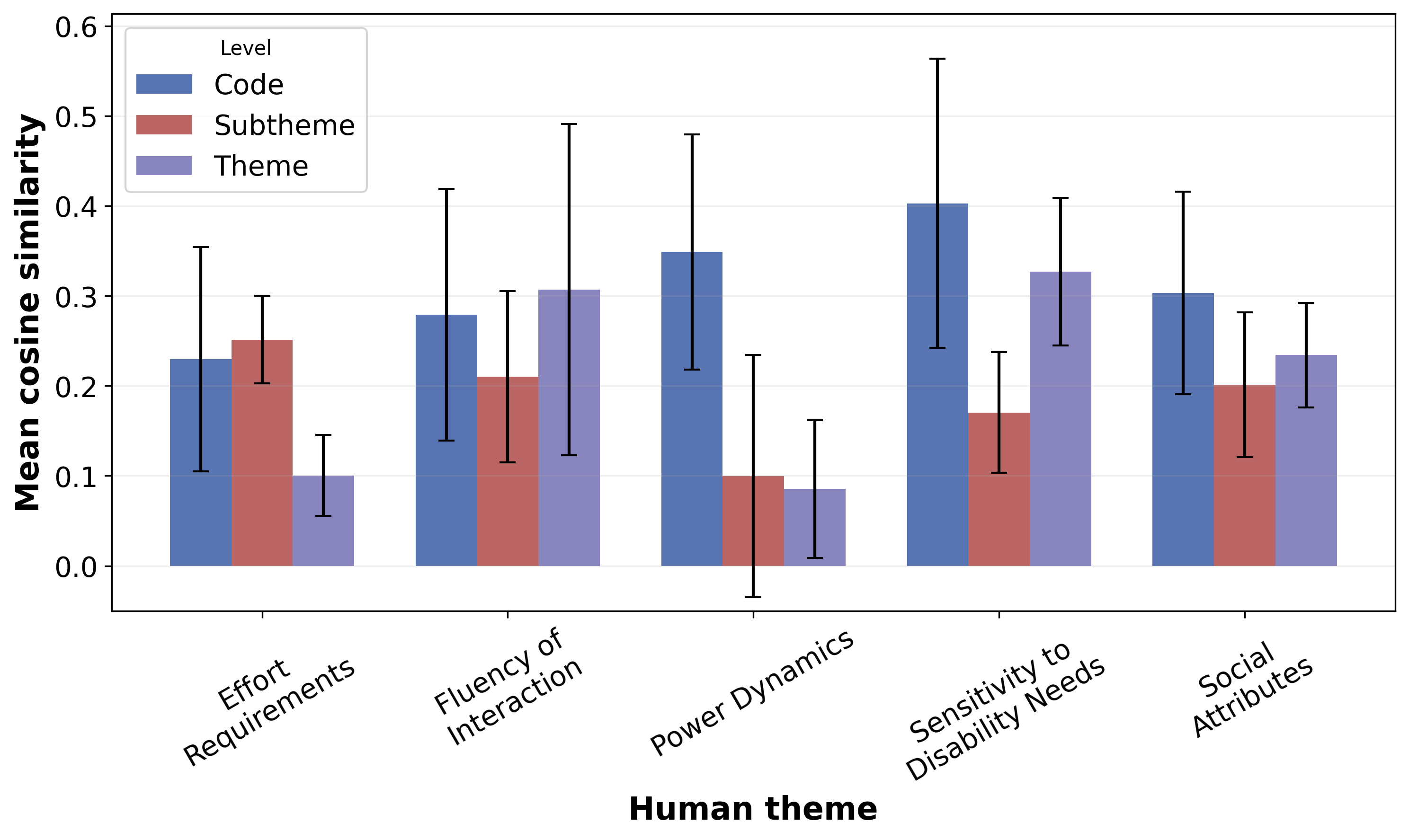}
    \vspace{-1em}
    \caption{Cosine similarity of human and LLM codes, subthemes, and themes averaged by human themes.}
    \label{RQ2}
    \vspace{-1.9em}
\end{figure}

\begin{table*}[t!]
\centering
\footnotesize
\caption{Qualitative comparison and evaluation of codes. The top 3 similar (S1, S2, S3) and dissimilar (D3, D2, D1) codes are presented based on their cosine similarity. IG: Interpretive Grounding, RF: Representational Fidelity, BOR: Bias/Marginalisation Risk.}
\vspace{-1em}
\label{tab}
\scriptsize
\begin{tabular}{p{8.3cm} p{1.5cm} p{2cm} p{0.9cm} p{0.7cm} p{0.7cm} p{0.7cm}}
\toprule
\textbf{Utterance}  & \textbf{Human Code} & \textbf{LLM Code} & \textbf{Cos sim.} &  \textbf{IG} & \textbf{RF} & \textbf{BMR} \\
\bottomrule

S1. ``I'm very aware of everything that we do, every every subject, including technology and sciences, are biased. Currently very, very biased so. I'm a bit worried that these robots might have certain. Unintentional agendas and viewpoints, like sexism, racism, and ableism.'' &  Concern over ableism and other biases & Concern about AI bias and discrimination & 0.639 &

Medium & Medium & High  \\\midrule

S2. ``The hand gestures were quite lifelike while they look like actually a human hand gestures. So that was quite nice to sort of experience'' &  Body language & Lifelike gestures positively noticed & 0.561 &

Medium & Medium & Low \\\midrule

S3. ``It was kind of weird to use it's hands to, like make up for it. And it was just. I've never seen something like move its hands so much and for someone who has anxiety that's quite like you just heard people by their hand like they're like actions in the way it just felt like very overwhelming.'' &  Anxiety could be exacerbated & Excessive hand movement overwhelming for anxious users & 0.538 &

Low & Low & High \\\midrule
 
\multicolumn{7}{c}{...}\\\midrule
D3. ``Now he's like a conversational agent. It provides you this information and. He listened to you like as a sounding board. It would be great for him to actually help me with something, maybe help me write an essay.'' &  Sense of being listened to & Robot useful for practical tasks like emailing & 0.085 &
Low & Low & High \\\midrule

D2. ``At first, when I was interacting with robots because it was very lifelike so I figured it was like OK, just like like a normal conversation with a real life person. So I guess I got, I took the time to develop what I wanted to ask the question, but then the robot didn't allow me to tell you to continue with the questions.'' & Lifelike & Robot interrupts before user finishes speaking & 0.065
 & Low & Low & Medium \\\midrule

D1. ``It also has a really nice like encouraging touch, [...] kind of coding it into this conversational language. So that initial aspect to it and just, like encouragement and sort of presenting information the way that's informative, but also like soothes anxiety somewhat and kind of like 'so these are the options', kinda breaking down.'' & Breaks it down & Robot's encouraging tone soothes anxiety & 0.042

& Low & Low & Medium \\

\bottomrule
\end{tabular}
\vspace{-2.5em}
\end{table*}

Semantic consistency analysis in Figure~\ref{RQ2} reveals how closely the meanings of human and LLM labels align for the same utterances. At the code level, the highest semantic consistency was observed for utterances associated with \textit{``sensitivity to disability needs''}, suggesting that both human and LLM annotations converge most strongly in describing functional or need-based aspects. At the subtheme level, the highest consistency was observed for \textit{``effort requirements''}, with closely comparable levels of similarity also found in \textit{``fluency of interaction''} and \textit{``social attributes''}. 
At the theme level, the strongest semantic alignment was found for \textit{``sensitivity to disability needs''} and \textit{``fluency of interaction''}, indicating that these higher-level concepts are more similarly conceptualized by both humans and the LLM. In contrast, the lowest semantic similarity was observed for the \textit{``power dynamics''} and \textit{``effort requirements''} categories, suggesting greater variability in how this theme is expressed or interpreted. These results indicate that while LLM-generated labels may differ lexically from human labels, they can also retain semantic alignment, particularly for functional and interaction-related constructs.

\subsection{RQ3: Qualitative Evaluation}

The results of the three-criteria qualitative comparison are presented in Table~\ref{tab}. We discuss selected ratings to illustrate the basis for evaluation.
Utterances~S1, S3, and D3 each received \textit{High} ratings for Bias and Marginalisation Risk, though through distinct mechanisms. In Utterance~S1, the LLM imports the term discrimination, which is absent from the utterance, while omitting ableism, which the participant named explicitly. Given that this study centres on disabled populations, this omission constitutes a substantive marginalisation risk. In Utterance~S3, the LLM shifts from the participant's first-person disclosure of lived anxiety to a generalised claim about anxious users, subsuming individual experience into a population-level characterisation the participant did not invoke. In Utterance~D3, the LLM reframes a reflection on relational experience as a functional utility statement, plausibly shaped by biases that prioritise task-oriented robot framings over affective ones, erasing the participant's primary concern. Utterances~D2 and D1 received \textit{Low} ratings for Interpretive Grounding. Both cases demonstrate the same failure: the LLM extracts a discrete, lexically salient element of the utterance rather than deriving a code that accounts for its overall meaning. Coding requires interpretive labelling, not paraphrasing; in both cases, the LLM produces the latter. Utterance~S2 received \textit{Medium} ratings for Interpretive Grounding and Representational Fidelity. The LLM code is not inaccurate but remains descriptive, failing to abstract to the conceptual label the human coder identified.

\section{Discussion}
Our findings show that LLM-assisted thematic analysis achieves partial alignment with human analysis, but with important limitations. Quantitatively, the model performs above chance in grouping and shows moderate semantic similarity. However, qualitative inspection reveals that disagreements are systematic rather than random, often involving shifts in interpretation, abstraction, and representation. These differences are especially visible in areas tied to lived experience and social meaning, which are central to HRI with vulnerable populations.

For RQ1, LLM clusterings are sensible and clearly better than random, as shown by the AMI scores in Figure~\ref{fig:RQ1}. This indicates that the model captures meaningful structural patterns in the data rather than assigning labels arbitrarily. However, Figure~\ref{fig:RQ1_1} shows uneven alignment at a finer level: conceptually diffuse themes, such as power dynamics, do not map consistently to a single LLM category and are instead spread across multiple categories.
This suggests that although LLMs have the potential to support thematic analysis in HRI, their reliability decreases as the level of abstraction and interpretive ambiguity increases. 
This both supports prior work showing reasonable agreement between human and LLM-generated themes~\cite{Castellanos_Jiang_Gomes_VanderMeer_Castillo_2025, Wen_Clough_Paton_Middleton_2026} and challenges it by demonstrating that agreement is not uniform across themes and can decrease for abstract and socially situated concepts.

RQ2 further reinforces this pattern by showing that semantic similarity is highest on average at the code level and decreases at the subtheme and theme levels, as presented in Figure~\ref{RQ2}. This is consistent with the idea that coding is closer to summarisation or surface-level interpretation, where LLMs are known to perform well. In contrast, generating subthemes and themes requires higher levels of abstraction, synthesis, and contextual reasoning, and LLM overall shows lower similarity at these levels. Interestingly, subthemes sometimes perform worse than themes. This likely reflects their position at an intermediate level of abstraction, where they must coherently group codes without the flexibility of broader themes. This likely reflects the model's difficulty at an intermediate level of abstraction, where it must coherently group codes beyond utterance-level description without the flexibility of broader themes. 
Overall, for HRI research, our findings imply that LLMs can be effectively leveraged for lower-level analytical tasks such as initial coding or data familiarisation, but are less reliable when ``zooming out'' to capture the holistic, relational, and contextually embedded aspects of participant experience.

For RQ3, the qualitative comparison reveals that LLM-generated codes systematically tend toward paraphrase and description rather than interpretive labelling. Coding is a theory-building act requiring immersion, abstraction, and progressive meaning-making~\cite{williams2019art}. As codes form the foundation of knowledge about human experience, systematically shallow coding produces impoverished knowledge, with particular consequences for vulnerable populations already at risk of misrepresentation. Representational fidelity compounds this concern. Several cases show LLMs cherry-picking the most lexically prominent element of an utterance, consistent with the risk that LLM interpretations are imposed rather than emergent~\cite{Schroeder_AubinRandazzo_Mimno_Schoenebeck_2025}. The substitution of the participant's term `ableism' with the broader `discrimination' in Table~\ref{tab}, for instance, is better understood as semantic sanitisation than hallucination: it introduces no unrelated content, but flattens a disability-specific account of harm into a generic category. This is in direct tension with qualitative traditions requiring meaning to arise bottom-up from data, and raises consent-related concerns about reframing participant perspectives. The bias and marginalisation findings point to a broader epistemic problem: LLMs may default to hegemonic framings or generalise individual experiences to population-level categories. Human researchers also produce situated interpretations, but bring disciplinary expertise, awareness of structural inequalities, and a relational accountability to participants that can partially counteract such tendencies~\cite{Schroeder_AubinRandazzo_Mimno_Schoenebeck_2025}. These tendencies were observed under a single model and may reflect its alignment training and default behaviours rather than LLM-assisted TA in general; other architectures may exhibit different coding and abstraction patterns, which cross-model comparison should examine.







\section{Conclusion}
\vspace{-0.5em}
We conducted a comparative analysis of human and LLM-generated TA for HRI related to vulnerable populations. Overall, our findings suggest a bounded role for LLM thematic analysis. They can support early-stage analysis (e.g., coding, clustering), but human-in-the-loop mechanisms are still valuable for higher-level interpretations. This is particularly critical for topics involving identity, power, and lived experience, where misrepresentations are not only more likely but can have ethical implications. Although this study focuses on disability, the observed patterns likely extend to other vulnerable populations. 
Future work should test this across contexts and examine hybrid human-AI workflows in which human researchers review and refine LLM-generated codes and themes to mitigate bias, overgeneralisation, and loss of participant meaning.\\

\noindent
\textbf{Contributions:} Conceptualisation: AM, FID, HG; Data curation: AM; Formal analysis: JB, FID, AM; Investigation: FID, AM; Methodology: AM, FID; Software: FID; Writing – original draft: AM, FID; Writing – review \& editing: HG, AM, FID; Funding acquisition: HG, AM; Supervision: HG; Visualisation: FID.

\bibliographystyle{IEEEtran}
\bibliography{references}

@article{Castellanos_Jiang_Gomes_VanderMeer_Castillo_2025, title={Large Language Models for Thematic Summarization in Qualitative Health Care Research: Comparative Analysis of Model and Human Performance}, ISSN={2817-1705}, DOI={10.2196/64447},  journal={Jmir Ai}, author={Castellanos, Arturo and et.al.}, year={2025}}

@article{Ornelas_Arajo_Ara_Trinkenreich_Kalinowski_2025, title={LLM-Assisted Thematic Analysis: Opportunities, Limitations, and Recommendations}, DOI={10.48550/arXiv.2511.14528}, abstractNote={[Context] Large Language Models (LLMs) are increasingly used to assist qualitative research in Software Engineering (SE), yet the methodological implications of this usage remain underexplored. Their integration into interpretive processes such as thematic analysis raises fundamental questions about rigor, transparency, and researcher agency. [Objective] This study investigates how experienced SE researchers conceptualize the opportunities, risks, and methodological implications of integrating LLMs into thematic analysis. [Method] A reflective workshop with 25 ISERN researchers guided participants through structured discussions of LLM-assisted open coding, theme generation, and theme reviewing, using color-coded canvases to document perceived opportunities, limitations, and recommendations. [Results] Participants recognized potential efficiency and scalability gains, but highlighted risks related to bias, contextual loss, reproducibility, and the rapid evolution of LLMs. They also emphasized the need for prompting literacy and continuous human oversight. [Conclusion] Findings portray LLMs as tools that can support, but not substitute, interpretive analysis. The study contributes to ongoing community reflections on how LLMs can responsibly enhance qualitative research in SE.}, journal={arXiv:2511.14528}, publisher={arXiv}, author={Ornelas, Tatiane and et. al.}, year={2025}}

@article{Veling_McGinn_2021, title={Qualitative Research in HRI: A Review and Taxonomy}, ISSN={1875-4805}, DOI={10.1007/s12369-020-00723-z}, abstractNote={The field of human–robot interaction (HRI) is young and highly inter-disciplinary, and the approaches, standards and methods proper to it are still in the process of negotiation. This paper reviews the use of qualitative methods and approaches in the HRI literature in order to contribute to the development of a foundation of approaches and methodologies for these new research areas. In total, 73 papers that use qualitative methods were systematically reviewed. The review reveals that there is widespread use of qualitative methods in HRI, but very different approaches to reporting on it, and high variance in the rigour with which the approaches are applied. We also identify the key qualitative methods used. A major contribution of this paper is a taxonomy categorizing qualitative research in HRI in two dimensions: by ’study type’ and based on the specific qualitative method used.}, journal={International Journal of Social Robotics}, author={Veling, Louise and McGinn, Conor}, year={2021}}

@article{Wen_Clough_Paton_Middleton_2026, title={Leveraging large language models for thematic analysis: a case study in the charity sector}, volume={41}, ISSN={1435-5655}, DOI={10.1007/s00146-025-02487-4}, abstractNote={This study explores how large language models (LLMs) can support deductive and inductive thematic coding in real-life contexts, balancing AI-driven efficiency with essential human oversight. Using three datasets from Tearfund, a UK-based Christian charity, we propose a dual-role human–LLM collaborative framework where the LLM functions as an initial annotator and a validator. In the deductive phase, GPT-4o and GPT-4o-mini were compared against human coders. GPT-4o achieved a substantial agreement in multi-label thematic categorization (κ = 0.61–0.65), while GPT-4o-mini showed a moderate agreement (κ = 0.41–0.58). Both models excelled in sentiment analysis (κ = 0.91–0.95), but struggled with evaluating evidence of impact due to contextual complexity (κ ≤ 0.01). GPT-4o-mini exhibited greater output variability and instability than GPT-4o, but benefited more from few-shot learning to mitigate hallucinations. In the inductive phase, GPT-4o demonstrated a strong semantic alignment with human-generated themes (cosine similarity = 0.76–0.79) though its tendency toward broad themes required human refinement. Despite their potential to streamline thematic analysis, LLMs also pose limitations and implementation challenges, including inconsistencies in excerpt extraction (precision = 0.41, recall = 0.53) and the trade-off between the time saved in coding and the time required for human validation. To facilitate practical implementation, we provide reusable prompt templates for four stages: context, instructions, data processing, and verification. Our findings underline the indispensable role of human expertise—from prompt engineering and managing hallucinations to final verification—to ensure accurate and trustworthy AI-assisted analyses. While LLMs can enhance qualitative analysis, their full potential is only realized under skilled human guidance.}, number={1}, journal={AI \& SOCIETY}, author={Wen, Chuanchi and Clough, Paul and Paton, Rachel and Middleton, Rebecca}, year={2026}, month=jan, pages={731–748}, language={en} }

@article{Montes_Feldt_Martos_Ouhbi_Premanandan_Graziotin_2025, title={Large Language Models in Thematic Analysis: Prompt Engineering, Evaluation, and Guidelines for Qualitative Software Engineering Research}, DOI={10.48550/arXiv.2510.18456}, abstractNote={As artificial intelligence advances, large language models (LLMs) are entering qualitative research workflows, yet no reproducible methods exist for integrating them into established approaches like thematic analysis (TA), one of the most common qualitative methods in software engineering research. Moreover, existing studies lack systematic evaluation of LLM-generated qualitative outputs against established quality criteria. We designed and iteratively refined prompts for Phases 2-5 of Braun and Clarke’s reflexive TA, then tested outputs from multiple LLMs against codes and themes produced by experienced researchers. Using 15 interviews on software engineers’ well-being, we conducted blind evaluations with four expert evaluators who applied rubrics derived directly from Braun and Clarke’s quality criteria. Evaluators preferred LLM-generated codes 61% of the time, finding them analytically useful for answering the research question. However, evaluators also identified limitations: LLMs fragmented data unnecessarily, missed latent interpretations, and sometimes produced themes with unclear boundaries. Our contributions are threefold. First, a reproducible approach integrating refined, documented prompts with an evaluation framework to operationalize Braun and Clarke’s reflexive TA. Second, an empirical comparison of LLM- and human-generated codes and themes in software engineering data. Third, guidelines for integrating LLMs into qualitative analysis while preserving methodological rigour, clarifying when and how LLMs can assist effectively and when human interpretation remains essential.}, journal={arXiv:2510.18456 [cs]}, publisher={arXiv}, author={Montes, Cristina Martinez and et.al.}, year={2025}}

@inproceedings{10.1145/3757279.3785593,
author = {Markelius, Alva and et.al.},
title = {Social Robotics for Disabled Students: An Empirical Investigation of Embodiment, Roles, and Interaction},
year = {2026},
doi = {10.1145/3757279.3785593},
booktitle = {ACM/IEEE International Conference on HRI},
}

@article{Mathis_Zhao_Pratt_Weleff_DePaoli_2024, title={Inductive thematic analysis of healthcare qualitative interviews using open-source large language models: How does it compare to traditional methods?}, ISSN={0169-2607}, DOI={10.1016/j.cmpb.2024.108356}, author={Mathis, Walter S and et.al.}, year={2024}, journal = {Computer Methods and Programs in Biomedicine}}

@article{Parkington_Teferra_RouleauTang_Perivolaris_Rueda_Dubrowski_Kapralos_Samavi_Greenshaw_Zhang_etal._2025, title={Human vs. LLM-Based Thematic Analysis for Digital Mental Health Research: Proof-of-Concept Comparative Study}, DOI={10.48550/arXiv.2507.08002}, journal={arXiv:2507.08002 [cs]},  author={Parkington, Karisa and et.al.}, year={2025}}

@article{Pascoe_2022, title={Reflections on a Systematic Literature Review: Questioning the (In)visibility of Researcher Positionality}, volume={46}, ISSN={1070-5309}, DOI={10.1093/swr/svac006}, number={2}, journal={Social Work Research}, author={Pascoe, Katheryn Margaret}, year={2022}, pages={176–180} }

@article{terry2017thematic,
  title={Thematic analysis},
  author={Terry, Gareth and Hayfield, Nikki and Clarke, Victoria and Braun, Virginia and others},
  journal={The SAGE handbook of qualitative research in psychology},
  year={2017},
  publisher={SAGE Publications Ltd}
}

@article{Braun_Clarke_2022, address={US}, title={Conceptual and design thinking for thematic analysis}, volume={9}, ISSN={2326-3598}, DOI={10.1037/qup0000196}, abstractNote={Thematic analysis (TA) is widely used in qualitative psychology. In using TA, researchers must choose between a diverse range of approaches that can differ considerably in their underlying (but often implicit) conceptualizations of qualitative research, meaningful knowledge production, and key constructs such as themes, as well as analytic procedures. This diversity within the method of TA is typically poorly understood and rarely acknowledged, resulting in the frequent publication of research lacking in design coherence. Furthermore, because TA offers researchers something closer to a method (a transtheoretical tool or technique) rather than a methodology (a theoretically informed framework for research), one with considerable theoretical and design flexibility, researchers need to engage in careful conceptual and design thinking to produce TA research with methodological integrity. In this article, we support researchers in their conceptual and design thinking for TA, and particularly for the reflexive approach we have developed, by guiding them through the conceptual underpinnings of different approaches to TA, and key design considerations. We outline our typology of three main “schools” of TA—coding reliability, codebook, and reflexive—and consider how these differ in their conceptual underpinnings, with a particular focus on the distinct characteristics of our reflexive approach. We discuss key areas of design—research questions, data collection, participant/data item selection strategy and criteria, ethics, and quality standards and practices—and end with guidance on reporting standards for reflexive TA. (PsycInfo Database Record (c) 2025 APA, all rights reserved)}, number={1}, journal={Qualitative Psychology}, publisher={Educational Publishing Foundation}, author={Braun, Virginia and Clarke, Victoria}, year={2022}, pages={3–26} }

@inproceedings{Schroeder_AubinRandazzo_Mimno_Schoenebeck_2025, title={Large Language Models in Qualitative Research: Uses, Tensions, and Intentions}, ISBN={979-8-4007-1394-1}, DOI={10.1145/3706598.3713120}, booktitle={CHI Conference on Human Factors in Computing Systems}, publisher={ACM}, author={Schroeder, Hope and et.al.}, year={2025}}

@article{williams2019art,
  title={The art of coding and thematic exploration in qualitative research},
  author={Williams, Michael and Moser, Tami},
  journal={International management review},
  year={2019}
}

@article{anthropic2025system,
  title={System card: Claude Sonnet 4.6},
  author={Anthropic, AI},
  journal={Claude-4 Model Card},
  year={2026}, 
  url={https://www.anthropic.com/system-cards}
}

@book{braun2021thematic,
  title={Thematic analysis: A practical guide},
  author={Braun, Virginia and Clarke, Victoria},
  year={2021},
  publisher={SAGE publications Ltd}
}

@article{Naeem_Smith_Thomas_2025, title={Thematic Analysis and Artificial Intelligence: A Step-by-Step Process for Using ChatGPT in Thematic Analysis},  ISSN={1609-4069}, DOI={10.1177/16094069251333886}, abstractNote={This study sets out how to use generative artificial intelligence (AI) in the six steps of systematic thematic analysis. It leverages AI to address the limitations of traditional thematic analysis. This paper developed prompts (inputs) for ChatGPT (a generative AI chatbot based on a large language model) that are based on many researchers’ discussions and criticisms of qualitative data analysis. The contributions of this paper are twofold. First, it addresses a critical research gap by showcasing ChatGPT prompts for each step of the six steps of systematic thematic analysis, which also addresses researcher training in thematic analysis. Second, it contributes to the development of input to train AI in thematic analysis, including a description of how to familiarize an AI system with the context of a research study and the researcher’s methodological and theoretical considerations; this approach helps to reduce human bias and improves accountability and transparency in thematic analysis.}, journal={International Journal of Qualitative Methods}, author={Naeem, Muhammad and Smith, Tracy and Thomas, Lorna}, year={2025}, language={EN} }

@article{Estvez_Terano_2021, title={A Case Study of a Robot-Assisted Speech Therapy for Children with Language Disorders},  ISSN={2071-1050}, DOI={10.3390/su13052771}, abstractNote={The aim of this study was to explore the potential of using a social robot in speech therapy interventions in children. A descriptive and explorative case study design was implemented involving the intervention for language disorder in five children with different needs with an age ranging from 9 to 12 years. Children participated in sessions with a NAO-type robot in individual sessions. Qualitative methods were used to collect data on aspects of viability, usefulness, barriers and facilitators for the child as well as for the therapist in order to obtain an indication of the effects on learning and the achievement of goals. The main results pointed out the affordances and possibilities of the use of a NAO robot in achieving speech therapy and educational goals. A NAO can contribute towards eliciting motivation, readiness towards learning and improving attention span of the children. The results of the study showed the potential that NAO has in therapy and education for children with different disabilities. More research is needed to gain insight into how a NAO can be applied best in speech therapy to make a more inclusive education conclusions.}, journal={Multidisciplinary Digital Publishing Institute}, author={Estévez, David and et.al.}, year={2021}}

@article{Keane_andWilliams_2019, title={Humanoid robots: learning a programming language to learn a traditional language}, volume={28}, ISSN={1475-939X}, DOI={10.1080/1475939X.2019.1670248}, abstractNote={This research is part of a larger three-year study investigating the impact of humanoid robots on students’ learning and engagement. In this case study, Aboriginal and non-Aboriginal students worked with a humanoid robot to develop, in parallel, both their programming skills and their understanding of the traditional Narungga language and culture. For six months a school engaged students in learning two languages: the coding language required to program the robot and the Narungga language. Qualitative data were collected and triangulated to determine how the humanoid robot was utilised in the classroom and re-occurring themes were identified through the case study. This research drew on questionnaires, interviews and journals from teachers to understand the impact of humanoid robots on student learning. The case study demonstrated how using humanoid robots enhanced pride and interest in Aboriginal language and culture.}, number={5}, journal={Technology, Pedagogy and Education}, publisher={Routledge}, author={Keane, Therese and et.al.}, year={2019}, month=oct, pages={533–546} }

@article{Moyle_Bramble_Jones_Murfield_2018, title={Care staff perceptions of a social robot called Paro and a look-alike Plush Toy: a descriptive qualitative approach}, volume={22}, ISSN={1360-7863}, DOI={10.1080/13607863.2016.1262820}, abstractNote={Objectives: Social robots such as Paro, a therapeutic companion robot, have recently been introduced into dementia care as a means to reduce behavioural and psychological symptoms of dementia. The purpose of this study was to explore care staff perceptions of Paro and a look-alike non-robotic animal, including benefits and limitations in dementia care. Methods: The study assumed a descriptive qualitative approach, nested within a large cluster-randomised controlled trial. We interviewed a subsample of 20 facility care staff, from nine long-term care facilities in Southeast Queensland, Australia. Thematic analysis of the data, which was inductive and data-driven, was undertaken with the assistance of the qualitative software, ATLAS.ti®. Results: The findings refer to four categories: increasing excitement for Paro and decreasing enthusiasm for Plush Toy; value and function of Paro; opportunities for engagement; and alternatives vs. robustness. Conclusion: Staff caring for people with dementia preferred Paro compared to a look-alike Plush Toy. Staff identified that Paro had the potential to improve quality of life for people with dementia, whereas the Plush Toy had limitations when compared to Paro. However, participants expressed concern that the cost of Paro could reduce opportunities for use within aged care.}, number={3}, journal={Aging \& Mental Health}, publisher={Routledge}, author={Moyle, Wendy and Bramble, Marguerite and Jones, Cindy and Murfield, Jenny}, year={2018}, month=mar, pages={330–335} }

@article{Page_Charteris_Berman_2021, title={Telepresence Robot Use for Children with Chronic Illness in Australian Schools: A Scoping Review and Thematic Analysis}, DOI={10.1007/s12369-020-00714-0}, journal={International Journal of Social Robotics}, author={Page, Angela and et.al.}, year={2021}}

@article{Poulsen_Burmeister_Greig_Ulhaq_Tien_2025, title={Value Sensitive Design of Social Robots: Enhancing the Lives of LGBT+ Older Adults}, ISSN={1875-4805}, DOI={10.1007/s12369-024-01201-6},  journal={International Journal of Social Robotics}, author={Poulsen, Adam and Burmeister, Oliver K. and Greig, Jenni and Ulhaq, Anwaar and Tien, David}, year={2025}}

@inproceedings{Simo_Avelino_Duarte_Figueiredo_2018, title={GeeBot: A Robotic Platform for Refugee Integration}, DOI={10.1145/3173386.3177833}, abstractNote={The refugee crisis is one of society’s leading challenges. After a journey for survival, refugees and host institutions face barriers that hinder the integration process. To design solutions, we interviewed two groups: host institutions and past refugees. We identified critical issues, from legal concerns, like unfamiliarity of their Refugee Status, to grocery shopping. Our envisioned solution is GeeBot, a low-cost egg-shaped robot that institutions would lend to arriving families for eighteen months. GeeBot will be a translator with teaching functions, an information provider, and an active promoter of interaction between native and refugee populations.}, booktitle={ ACM/IEEE International Conference on HRI},  author={Simão, Hugo and Avelino, João and Duarte, Nuno and Figueiredo, Rui}, year={2018}}

@article{Sraa_Tndel_Kharas_Serrano_2023, title={What do Older Adults Want from Social Robots? A Qualitative Research Approach to Human-Robot Interaction (HRI) Studies}, volume={15}, ISSN={1875-4805}, DOI={10.1007/s12369-022-00914-w}, abstractNote={This study investigates what older adults want from social robots. Older adults are often presented with social robots designed based on developers’ assumptions that only vaguely address their actual needs. By lacking an understanding of older adults’ opinions of what technology should or could do for them–and what it should not do–we risk users of robots not finding them useful. Social and humanistic research on the robotization of care argues that it is important to prioritize user needs in technology design and implementation. Following this urgent call, we investigate older adults’ experiences of and approach to social robots in their everyday lives. This is done empirically through a qualitative analysis of data collected from six group interviews on care robots with health care service users, informal caregivers (relatives), and professional caregivers (healthcare workers). Through this “Need-Driven-Innovation” study we argue that, to secure a functional and valuable technology-fit for the user, it is crucial to take older adults’ wishes, fears, and desires about technology into account when implementing robots. It is also crucial to consider their wider networks of care, as the people in these networks also often interact with the assistive technology service users receive. Our study shows that more qualitative knowledge on the social aspect of human-robot interaction is needed to support future robot development and use in the health and care field and advocates for the crucial importance of strengthening the position of user-centered qualitative research in the field of social robotics.}, number={3}, journal={International Journal of Social Robotics}, author={Søraa, Roger Andre and Tøndel, Gunhild and Kharas, Mark W. and Serrano, J Artur}, year={2023}, month=mar, pages={411–424}, language={en} }

@inproceedings{Tozadore_Guneysui_2023, title={Teacher’s Perception on Social Robots to Promote the Integration of Children with Migration Background}, ISBN={979-8-4007-0824-4},  DOI={10.1145/3623809.3623937}, booktitle={International Conference on Human-Agent Interaction}, publisher={ACM}, author={Tozadore, Daniel C. and Guneysu Ozgur, Arzu and Kuoppamäki, Sanna}, year={2023}}

@article{Williams_Townsend_Naeche_Chapman-Nisar_Hollis_Slovak_Minds_2023, title={Investigating the Feasibility, Acceptability, and Appropriation of a Socially Assistive Robot Among Minority Youth at Risk of Self-Harm: Results of 2 Mixed Methods Pilot Studies}, DOI={10.2196/52336}, journal={JMIR Formative Research},  author={Williams, A. Jess and et.al.}, year={2023}}

@article{JMLR:v11:vinh10a,
  author  = {Nguyen Xuan Vinh and Julien Epps and James Bailey},
  title   = {Information Theoretic Measures for Clusterings Comparison: Variants, Properties, Normalization and Correction for Chance},
  journal = {Journal of Machine Learning Research},
  year    = {2010},
}

\end{document}